\documentclass[11pt,a4paper,dvipsnames]{article}
\usepackage[hyperref]{acl2021}
\usepackage{times}
\usepackage{latexsym}

\usepackage{amsmath}
\DeclareMathOperator*{\argmax}{argmax}
\usepackage{graphics}
\usepackage[T1]{fontenc}
\usepackage{array}
\usepackage{arydshln}
\usepackage{amssymb}
\usepackage{graphicx}
\usepackage{multirow}
\usepackage{multicol}
\usepackage{tabularx}
\usepackage{booktabs}
\usepackage{xcolor}

\definecolor{green(munsell)}{rgb}{0.0, 0.66, 0.47}

\usepackage{microtype}
\aclfinalcopy 

\title{RewardsOfSum: Exploring Reinforcement Learning Rewards\\for Summarisation}

\author{Jacob Parnell\textsuperscript{1,2}, Inigo Jauregi Unanue\textsuperscript{1,2}, Massimo Piccardi\textsuperscript{1} \\
  \textsuperscript{1}University of Technology Sydney, NSW, Australia \\
  \textsuperscript{2}RoZetta Technology, NSW, Australia \\
  \{\texttt{jacob.parnell,inigo.jauregi}\}\texttt{@rozettatechnology.com} \\
  \texttt{massimo.piccardi@uts.edu.au} \\}

\date{}

\begin{document}

\maketitle
\begin{abstract}
To date, most abstractive summarisation models have relied on variants of the negative log-likelihood (NLL) as their training objective. In some cases, reinforcement learning has been added to train the models with an objective that is closer to their evaluation measures (e.g. ROUGE). However, the reward function to be used within the reinforcement learning approach can play a key role for performance and is still partially unexplored. For this reason, in this paper, we propose two reward functions for the task of abstractive summarisation: the first function, referred to as \textit{RwB-Hinge}, dynamically selects the samples for the gradient update. The second function, nicknamed \textit{RISK}, leverages a small pool of strong candidates to inform the reward. In the experiments, we probe the proposed approach by fine-tuning an NLL pre-trained model over nine summarisation datasets of diverse size and nature. The experimental results show a consistent improvement over the negative log-likelihood baselines.
\end{abstract}

\section{Introduction}
\label{sec:Introduction}
The current state-of-the-art neural text summarisation models have been refined to excel at either the extractive or abstractive styles, or even both \citep{zhang2019, lewis2020, raffel2020exploring}. Along with contemporary summarisation datasets \citep{narayan2018dont, grusky2018, fabbri2019}, the advent of large pre-trained language models, and their subsequent derivations \citep{liu2019b, park2020continual}, has allowed summarisation to become a more practical and reasonable task to implement, without compromising, and often improving, the accuracy. However, these models usually employ the standard negative log-likelihood (NLL) as their training objective, which aims to maximise the likelihood of each token in a given ground-truth reference. Despite its efficacy, the NLL fails to account for synonymous tokens and other potentially valid variations, and strongly biases the model towards the ground-truth reference \citep{ranzato2016sequence}. Furthermore, the NLL operates as a token-level objective during training, which promotes an inconsistent comparison with sequence-level evaluation metrics, such as ROUGE \citep{lin2004}.

In order to address the inconsistency between token-level training and sequence-level evaluation, reinforcement learning (RL) has been adopted in summarisation and other language generation tasks to afford the optimization of sequence-level metrics during training \citep{paulus2017deep, pasunuru2018multireward}. Reinforcement learning has proved successful at improving the accuracy of language generation tasks, such as summarisation \citep{paulus2017deep, arumae2018reinforced, pasunuru2018multireward} and machine translation \citep{ranzato2016sequence, edunov2018classical}. However, balancing exploration and exploitation remains imperative to the successful choice of an effective reward. When standard RL techniques, such as REINFORCE \citep{williams92reinforce}, are implemented in natural language generation tasks, the required expectation becomes intractable due to large vocabulary sizes. Therefore, the application of REINFORCE is typically reduced to calculating the approximate expectation with respect to only a single predicted sequence. To teach the model to understand the importance of sample variation among synonymous tokens, we instead choose to implement an objective function which includes multiple predicted sequences, allowing for a scenario in which several valid candidate summaries can be considered. Another consideration is that the success of techniques such as REINFORCE strongly depends on the use of an effective and appropriate reward. Designing such a reward, one which enables the model to manipulate multiple sequences and yet provides a positive and informative outcome in the process, is therefore necessary for producing better results. This allows us to modify the reinforcement learning framework in such a way that enforces only a higher weighting to those predicted sequences which obtain a higher reward. As such, we apply two techniques to summarisation; \textit{RwB-Hinge}, which applies a hinge-loss modification to the classical REINFORCE with baseline \citep{rennie2017selfcritical} to selectively apply the model gradients, and \textit{Expected Risk Minimization (RISK)} \citep{edunov2018classical}, which leverages a small pool of strong sampled candidates to smartly inform the reward function.
We aptly refer to our framework as \textit{RewardsOfSum}, to hint at the exploration of suitable reward functions 
for summarisation. Empirically, we show that the two proposed variants 
perform better than standard negative log-likelihood baselines over a range of datasets of diverse size and nature.

\section{Related Work}
\label{sec:Related}
In recent years, there has been some work in summarisation to separate from the traditional negative log-likelihood (NLL) objective function, and mollify its dependency on ground-truth references. Several implementations of reinforcement learning in summarisation involved optimizing discrete metrics, such as the standard ROUGE \citep{paulus2017deep, narayan2018ranking}. Others have introduced novel rewards into the reinforcement learning framework, such as question-focused rewards \citep{arumae2018reinforced}, saliency and entailment rewards \citep{pasunuru2018multireward}, and even distributional semantic rewards \citep{li2019deep}. \citet{gao2020} also present a novel unsupervised metric for summarisation which correlates highly with discrete evaluation metrics if adopted in a reinforcement learning approach.

On the other hand, there has been much work in leveraging large, pre-trained language models (LM) \citep{devlin2018, lewis2020, raffel2020exploring} to improve the quality and performance of summarisation models. Utilizing pre-trained language models requires significantly less engineering effort to continually improve over state-of-the-art baselines. Typically, these approaches include using novel pre-training objectives \citep{zhang2019, raffel2020exploring, zhu2020make} or implementing successful reinforcement learning techniques \citep{bae2019summary}. \citet{li2019deep} found that optimizing semantic rewards in reinforcement learning, using BERTScore \citep{zhang2020bertscore}, does not necessarily correlate with the ROUGE score at test time. As such, the choice of reward in a reinforcement learning approach should attempt to carefully align with the evaluation metric.

How best to inform the reward via the reward function, is critical to the performance of models in an RL framework. In our work, we aim to stray from the typical sole NLL objective, and by leveraging a pre-trained language model in a reinforcement learning framework, explore different RL-based reward functions for summarisation.

\section{Proposed Reinforcement Learning Training}
\label{sec:RL Training}
In order to improve over the negative log-likelihood baseline models, we aim to implement a reinforcement learning framework that adopts the standard evaluation metric, ROUGE, as a reward during training. We aim to keep consistent with previous implementations of reinforcement learning in summarisation, and assume ROUGE-L F1 to be the reward metric in the following work.

In Sections \ref{subsec:rwbhinge} and \ref{subsec:riskloss}, we consider the following standard notations: $x$ is defined as an input source document, $y^{*}$, $\hat{y}$, and $y^{s}$ are referred to as the ground-truth reference, argmax prediction, and sampled sequence, respectively, and $r(y)$ refers to the reward of sequence $y$, computed with respect to the ground-truth reference, $y^*$. By exploiting a combination of sampling and predictions, we aim to enhance training diversity in the vein of the work of \citet{lijurafsky2016, lijurafsky2016_2, holtzmann2020}.

\subsection{RwB-Hinge}
\label{subsec:rwbhinge}
We adopt the standard self-critical policy gradient objective \citep{rennie2017selfcritical}, notably applied to summarisation by \citet{paulus2017deep}:

\begin{equation}
\label{eq:RwB_hinge1}
\begin{aligned}
\alpha = -[r(y^{s}) - r(\hat{y})] \\
\end{aligned}
\end{equation}

\vspace{-18pt}

\begin{equation}
\label{eq:RwB_hinge2}
\begin{aligned}
L_{RwB} = \alpha \sum\limits_{t=1}^{n^{'}} \: \log p(y_{t}^{s}|y_{1},\dots,y_{t-1}, x) \\
%
\end{aligned}
\end{equation}

In (\ref{eq:RwB_hinge1}), $y^s$ and $\hat{y}$ denote a sampled sequence and the argmax prediction of the current model, respectively. The reward of the argmax, $r(\hat{y})$, is used as a ``baseline'' for the reward of the sample, $r(y^s)$. It is easy to see that if $r(y^s) - r(\hat{y}) > 0$, the sign of this loss is negative, treating $y^s$ as a ``good'' prediction and leading to an increase of its probability. Conversely, if the sign is positive, $y^s$ is deemed as a ``bad'' prediction and its probability is decreased.

However, in abstractive summarisation it is not trivial to discriminate between a good and a bad summary when the reward score is in an intermediate range. To avoid inappropriately penalising acceptable predictions, we propose incorporating a hinge loss in (\ref{eq:RwB_hinge1}): 

\begin{equation}
\label{eq:RwB_hinge3}
\begin{aligned}
\alpha = -\max{[0, (r(y^{s}) - r(\hat{y}))]} \\
\end{aligned}
\end{equation}

The hinge loss allows the model to limit the gradient updates to only the predictions that are considered as good. In this way, we avoid the risk of unstable training updates and hope to  afford a clearer trajectory towards a well-trained model.

\subsection{Expected RISK Minimization}
\label{subsec:riskloss}
We also utilise a classical structured loss function that has been shown to perform well in sequence-to-sequence learning tasks \citep{edunov2018classical}:

\begin{equation}
\label{eq:RISK}
\begin{aligned}
L_{RISK} = \sum\limits_{y \in U(x)}^{} -\textit{r}(y) \cdot \textit{p}(y|x, \theta) \\
\end{aligned}
\end{equation}

In (\ref{eq:RISK}), $y$ represents one of multiple candidate summaries, sampled or predicted with the methods defined in Section \ref{subsec:sampling} (e.g. argmax, Gumbel-Softmax \citep{jang2017categorical}), that form the total candidate summary set $U(x)$. The conditional probability of the predicted summary is noted as $p(y|x, \theta)$.

This conditional probability is defined in (\ref{eq:RISK_supp}), where $m$ is the number of tokens in the summary. The sum of logarithms in (\ref{eq:RISK_supp2}) is divided by the total number of tokens in the sequence, and is scaled back using an exponential function, allowing each candidate summary to be compared fairly in the objective function and avoiding underflow.

\begin{equation}
\label{eq:RISK_supp}
\begin{aligned}
    \textit{p}(y|x, \theta) = \frac{f(y, x, \theta)}{\sum\limits_{y^{'} \in U(x)}^{} f(y^{'}, x,  \theta)} \\
\end{aligned}
\end{equation}

\begin{equation}
\label{eq:RISK_supp2}
\begin{aligned}
\eta = \sum\limits_{j=1}^{m} \: \textrm{log}p(u^{j}|u^{1},\dots,u^{j-1}, x, \theta) \\
\end{aligned}
\end{equation}
\begin{equation}
\label{eq:RISK_supp3}
\begin{aligned}
f(y, x, \theta) = \textrm{exp}[\frac{\eta}{m}] \\
\end{aligned}
\end{equation}

By using this objective function, the model is taught to assign higher probability to the candidate summaries that obtain higher rewards. This objective does not require a baseline or hinge loss to select the predictions, since using multiple candidates already 
exposes the model to different, potentially valid predictions. \citet{edunov2018classical} demonstrates the effectiveness of this approach at sentence level for both neural machine translation and summarisation. For the summarisation task, \citet{edunov2018classical} compute the reward at sentence-level since their dataset has single-sentence references. However, as the reward function is agnostic to single or multi-sentence predictions, we can easily translate the \textit{RISK} objective function to be used at summary level.

\subsection{Overall Training Objective}
\label{subsec:train_obj}
Similar to previous reinforcement learning implementations \citep{paulus2017deep, li2019deep}, we, too, utilise a mixed learning objective function, as shown in (\ref{eq:mixed1}). This mixed approach helps the model to not deviate too much from the reference summaries, given a $\gamma$ balancing coefficient chosen with a strict validation criterion (Appendix \ref{app:val_scores}). The $L_{RL}$ term refers to either the RwB-Hinge or RISK training objective function.

\begin{equation}
\label{eq:mixed1}
\begin{aligned}
L_{mixed} = \gamma L_{XENT} + (1 - \gamma) L_{RL} \\
\end{aligned}
\end{equation}

\vspace{6pt}

\section{Experimental Setup}
\label{sec:Experiments}

\subsection{Datasets}
\label{subsec:data}
Inspired by the recent work from \citet{zhang2019}, we utilise nine of the summarisation datasets reported in their paper. The nine datasets have been chosen based on the different lengths of their  reference summaries, to provide enough of a variation to demonstrate the applicability of the presented methods. We split the datasets into three classes: ``short'', ``medium'', and ``long''. Short datasets have reference summaries $\leq$ 64 tokens, medium datasets  $>$ 64 and $\leq$ 128 tokens, and long datasets $>$ 128 tokens.

\begin{table}[!ht]
\centering
\begin{tabular}{|c|ccc|}
\hline
\textbf{Dataset} & \textbf{Train} & \textbf{Test} & \textbf{Dev} \\
\hline
\textbf{AESLC} & 14.4K & 1.9K & 1.9K \\
\textbf{Gigaword} & 3.8M & 1.9K & 189K \\
\textbf{XSum} & 203K & 11.3K & 11.3K \\
\hline
\textbf{CNN/DM} & 287K & 11.4K & 13.3K \\
\textbf{Reddit-TIFU} & 33.7K & 4.2K & 4.2K \\
\textbf{Newsroom} & 995K & 108K & 108K \\
\hline
\textbf{Pubmed} & 119K & 6.6K & 6.6K \\
\textbf{ArXiv} & 203K & 6.4K & 6.4K \\
\textbf{Billsum} & 18.9K & 3.2K & 1.2K \\
\hline
\end{tabular}
\caption{Statistics on the datasets used in the experiments. Figures are rounded. The top third are short datasets ($\leq$ 64 tokens references summaries), the middle third are medium datasets ($>$ 64 and $\leq$ 128 tokens), and the bottom third are long datasets ($>$ 128 tokens).}
\label{tab:dataset_statistics}
\end{table}

\subsection{Sampling Methods}
\label{subsec:sampling}

In order to promote exploration across the vocabulary distribution, we employ three simple methodologies to provide candidate sequences for our training objectives.

\textbf{Argmax:} As is the standard with the majority of sequence generation tasks, a predicted sentence can be easily provided by allowing the model to make hard decisions (e.g. argmax) over the probability distribution generated by the decoder. This allows us to use it as a baseline for the following experiments. In its simplest form the argmax is defined as:

\begin{equation}
\label{eq:argmax}
\begin{aligned}
\hat{y}_{j} = \argmax\limits_{y} p(y|x,y^*_{j-1}, \theta) \quad j=1,\dots,n \\
\end{aligned}
\end{equation}

\noindent where we use ``teacher forcing'' for the predictions.

\textbf{2nd-Best:} Similar to the argmax, we employ a $k$-best approach to sample the second best-argmax from the same probability distribution generated by the decoder. This allows us to choose different, yet similarly weighted words from the decoder to introduce variability between produced summaries:

\begin{equation}
\label{eq:secondargmax}
\begin{aligned}
y^{s}_{j} = \argmax\limits_{k=2} p(y|x,y^*_{j-1}, \theta) \quad j=1,\dots,n \\
\end{aligned}
\end{equation}

\textbf{Gumbel-Softmax:} We also utilise a recent re-parameterization technique known as the Gumbel-Softmax \citep{jang2017categorical} that allows sampling soft latent categorical variables by transforming samples from a Gumbel distribution. Compared to the standard ``hard'' predictions, this approach is differentiable  and allows controlling the sparsity of the samples by a temperature parameter, $\tau$:

\begin{equation}
\label{eq:gumbel2}
\begin{aligned}
\tilde{p}^{i}_{j} = \frac{\textrm{exp}((\textrm{log}(p^{i}_{j}) + g^{i}) / \tau}{\sum_{v=1}^{V}\textrm{exp}((\textrm{log}(p^{v}_{j}) + g^{v}) / \tau} \\
\end{aligned}
\end{equation}

In (\ref{eq:gumbel2}), $g^{i}$ is a sample from the zero-mean, unit-scale Gumbel distribution, $p^{i}_{j}$ is the probability distribution for a given token $i$ at slot $j$, and the temperature parameter, $\tau$, controls the sparsity of the output soft variable, $\tilde{p}^{i}_{j}$. In our experiments, we have set $\tau$ to 0.1 to enforce sparsity.

\subsection{Baseline Model and Training Runs}
\label{subsec:baseline}
The abstractive text summarisation model we use for our experiments is PEGASUS, a large pre-trained Transformer encoder-decoder architecture that has recently reported state-of-the-art results over a number of datasets. Please refer to \citet{zhang2019} for details. All hyperparameters used in our experiments can be found in Appendix \ref{app:hyperparameters}.

We employ two training approaches to test the solidity of the proposed methods. The first is a few-shot learning approach that adopts limited, fixed numbers of training samples (1000) and training iterations (2000) for fine-tuning the model. The second is a full-data learning approach, that utilises all available training data, and exhausts the objective function until convergence over the validation set. In all experiments, we first fine-tune a pre-trained PEGASUS model with the NLL, and then we further fine-tune the NLL model with one of the proposed approaches. We train the model in this way to avoid the slow and inefficient training often associated with policy gradient objectives, and as a result, adhere to the standard warm-start NLL training adopted in previous reinforcement learning-based approaches \citep{paulus2017deep,li2019deep}.

In the following experiments, we refer to PEGASUS as PEG, and its NLL-tuned models with the suffixes -few\_shot and -full\_data. The proposed approaches are in turn noted as \textit{RwB-Hinge} and \textit{RISK}.

\begin{table}[!ht]
\centering
\begin{tabular}{|c|ccc|}
\hline
\textbf{Experiment} & \textbf{Arg-max} & \textbf{2nd-Best} & \textbf{G-S} \\
\hline
\textit{RwB-Hinge} & \checkmark & & \checkmark \\
\textit{RISK-2} & \checkmark & & \checkmark \\
\textit{RISK-3} & \checkmark & \checkmark & \checkmark \\
\hline
\end{tabular}
\caption{Different experiments and the different sampling methods used in each. Here, \textit{RISK-2} and \textit{RISK-3} denote the number of samples we utilise in the RISK objective function; two and three, respectively.}
\label{tab:experiments}
\end{table}

\begin{table*}[!ht]
\centering
\resizebox{\textwidth}{!}{%
\begin{tabular}{|c|ccc|ccc|ccc|}
\hline
\multirow{2}{30pt}{\textbf{Model}} & & \textbf{AESLC} & & & \textbf{Gigaword} & & & \textbf{XSum} & \\
 & \textbf{R-1} & \textbf{R-2} & \textbf{R-L} & \textbf{R-1} & \textbf{R-2} & \textbf{R-L} & \textbf{R-1} & \textbf{R-2} & \textbf{R-L} \\
\hline
PEG$_\textrm{few\_shot}$ & \textbf{29.96} & \textbf{14.54} & \textbf{29.17} & 31.81 & 13.19 & 29.12 & 41.81 & 18.32 & 33.50 \\
\hdashline
+ \textit{RwB-Hinge} & 28.69 & 13.83 & 27.82 & 31.83 & 13.15 & 29.08 & 42.47{$^\dagger$} & 18.82{$^\dagger$} & 33.94 \\
+ \textit{RISK-2} & 29.35 & 14.14 & 28.39 & 31.96 & 13.22 & 29.27 & 42.57{$^\dagger$} & 18.71{$^\dagger$} & 33.96{$^\dagger$} \\
+ \textit{RISK-3} & 29.28 & 14.05 & 28.31 & \textbf{32.10}{$^\dagger$} & \textbf{13.35}{$^\dagger$} & \textbf{29.43}{$^\dagger$} & \textbf{42.66}{$^\dagger$} & \textbf{19.01}{$^\dagger$} & \textbf{34.15}{$^\dagger$} \\
\hline
\hline
PEG$_\textrm{full\_data}$ & 32.63 & 15.84 & 32.19 & 33.81 & 14.26 & 30.89 & 41.52 & 18.21 & 33.31 \\
\hdashline
+ \textit{RwB-Hinge} & \textbf{34.39}{$^\dagger$} & \textbf{17.58}{$^\dagger$} & \textbf{33.71}{$^\dagger$} & \textbf{34.10}{$^\dagger$} & \textbf{14.52} & \textbf{31.31}{$^\dagger$} & 42.87{$^\dagger$} & \textbf{19.36} & 34.56{$^\dagger$} \\
+ \textit{RISK-2} & 33.55{$^\dagger$} & 17.01{$^\dagger$} & 32.91{$^\dagger$} & 33.97 & 14.45 & 31.18{$^\dagger$} & \textbf{42.93}{$^\dagger$} & 19.25{$^\dagger$} & \textbf{34.67}{$^\dagger$} \\
+ \textit{RISK-3} & 33.75{$^\dagger$} & 17.03{$^\dagger$} & 33.04{$^\dagger$} & 33.97 & \textbf{14.52} & 31.14{$^\dagger$} & 42.74{$^\dagger$} & 19.23{$^\dagger$} & 34.60{$^\dagger$} \\
\hline
\end{tabular}%
}
\caption{Results on short datasets: AESLC, Gigaword, and XSum. Here we compare the limited resource (PEG$_\textrm{few\_shot}$) and full-data (PEG$_\textrm{full\_data}$) approaches with our different implementations. ($\dagger$) means that the differences are statistically significant with respect to the baseline with a p-value < 0.05 over a bootstrap hypothesis test. Best ROUGE-1/2/L scores are bolded.}
\label{tab:improvements - small datasets}
\end{table*}

\begin{table*}[!ht]
\centering
\resizebox{\textwidth}{!}{%
\begin{tabular}{|c|ccc|ccc|ccc|}
\hline
\multirow{2}{30pt}{\textbf{Model}} & & \textbf{CNN/DM} & & & \textbf{Reddit-TIFU} & & & \textbf{Newsroom} & \\
 & \textbf{R-1} & \textbf{R-2} & \textbf{R-L} & \textbf{R-1} & \textbf{R-2} & \textbf{R-L} & \textbf{R-1} & \textbf{R-2} & \textbf{R-L} \\
\hline
PEG$_\textrm{few\_shot}$ & 40.65 & 17.60 & 37.81 & 24.84 & 7.21 & 20.12 & 33.33 & 20.01 & 29.17 \\
\hdashline
+ \textit{RwB-Hinge} & 40.44 & 17.44 & 37.54 & 25.55{$^\dagger$} & 7.23 & 20.09 & 34.03{$^\dagger$} & 20.74{$^\dagger$} & 29.86{$^\dagger$} \\
+ \textit{RISK-2} & 40.52 & 17.48 & 37.62 & 25.69{$^\dagger$} & 7.25 & 20.26 & 34.26{$^\dagger$} & 21.10{$^\dagger$} & 30.14{$^\dagger$} \\
+ \textit{RISK-3} & \textbf{40.76} & \textbf{17.63} & \textbf{37.87} & \textbf{25.73}{$^\dagger$} & \textbf{7.30} & \textbf{20.35} & \textbf{34.40}{$^\dagger$} & \textbf{21.27}{$^\dagger$} & \textbf{30.21}{$^\dagger$} \\
\hline
\hline
PEG$_\textrm{full\_data}$ & 40.58 & \textbf{18.15} & 37.94 & 23.66 & 6.72 & 19.24 & 36.39 & 23.90 & 32.50 \\
\hdashline
+ \textit{RwB-Hinge} & 40.84{$^\dagger$} & 17.74 & 38.19{$^\dagger$} & 23.95{$^\dagger$} & 6.93 & 19.69{$^\dagger$} & \textbf{36.85}{$^\dagger$} & \textbf{24.01} & \textbf{33.00}{$^\dagger$} \\
+ \textit{RISK-2} & \textbf{40.88}{$^\dagger$} & 17.91 & 38.19{$^\dagger$} & 24.25{$^\dagger$} & 7.19{$^\dagger$} & 20.00{$^\dagger$} & 36.74 & \textbf{24.01} & 32.73 \\
+ \textit{RISK-3} & \textbf{40.88}{$^\dagger$} & 17.91 & \textbf{38.28}{$^\dagger$} & \textbf{24.70}{$^\dagger$} & \textbf{7.46}{$^\dagger$} & \textbf{20.25}{$^\dagger$} & 36.04 & 23.22 & 32.18 \\
\hline
\end{tabular}%
}
\caption{Results on medium datasets: CNN/DM, Reddit-TIFU, and Newsroom. Here we compare the limited resource (PEG$_\textrm{few\_shot}$) and full-data (PEG$_\textrm{full\_data}$) approaches with our different implementations. ($\dagger$) means that the differences are statistically significant with respect to the baseline with a p-value < 0.05 over a bootstrap hypothesis test. Best ROUGE-1/2/L scores are bolded.}
\label{tab:improvements - medium datasets}
\end{table*}

\begin{table*}[!ht]
\centering
\resizebox{\textwidth}{!}{%
\begin{tabular}{|c|ccc|ccc|ccc|}
\hline
\multirow{2}{30pt}{\textbf{Model}} & & \textbf{Pubmed} & & & \textbf{ArXiv} & & & \textbf{Billsum} & \\
 & \textbf{R-1} & \textbf{R-2} & \textbf{R-L} & \textbf{R-1} & \textbf{R-2} & \textbf{R-L} & \textbf{R-1} & \textbf{R-2} & \textbf{R-L} \\
\hline
PEG$_\textrm{few\_shot}$ & 38.28 & 13.70 & 23.32 & 38.08 & 11.61 & 22.87 & 48.27 & 27.79 & 35.70 \\
\hdashline
+ \textit{RwB-Hinge} & 40.11{$^\dagger$} & 14.45{$^\dagger$} & 23.88{$^\dagger$} & 38.85{$^\dagger$} & 11.90{$^\dagger$} & 22.88 & 48.61{$^\dagger$} & \textbf{29.35}{$^\dagger$} & \textbf{36.91}{$^\dagger$} \\
+ \textit{RISK-2} & \textbf{40.19}{$^\dagger$} & \textbf{14.61}{$^\dagger$} & \textbf{23.98}{$^\dagger$} & \textbf{38.98}{$^\dagger$} & \textbf{12.02}{$^\dagger$} & \textbf{22.90} & 48.21 & 28.34{$^\dagger$} & 35.97 \\
+ \textit{RISK-3} & \textbf{40.19}{$^\dagger$} & 14.55{$^\dagger$} & 23.95{$^\dagger$} & 38.68{$^\dagger$} & 11.88{$^\dagger$} & 22.81 & \textbf{48.65} & 28.71{$^\dagger$} & 36.37{$^\dagger$} \\
\hline
\hline
PEG$_\textrm{full\_data}$ & 40.57 & 16.05 & \textbf{25.46} & 38.48 & 13.33 & 24.12 & 52.98 & 34.44 & 41.36 \\
\hdashline
+ \textit{RwB-Hinge} & \textbf{40.80} & \textbf{16.27} & 25.41 & \textbf{38.95}{$^\dagger$} & \textbf{13.69}{$^\dagger$} & \textbf{24.19} & \textbf{54.30}{$^\dagger$} & \textbf{36.01}{$^\dagger$} & \textbf{42.76}{$^\dagger$} \\
+ \textit{RISK-2} & 40.32 & 15.85 & 25.31 & 38.76 & 13.55 & 24.11 & 53.76{$^\dagger$} & 35.54{$^\dagger$} & 42.37{$^\dagger$} \\
+ \textit{RISK-3} & 40.36 & 15.89 & 25.26 & 38.42 & 13.37 & 24.12 & 54.27{$^\dagger$} & 35.80{$^\dagger$} & 42.51{$^\dagger$} \\
\hline
\end{tabular}%
}
\caption{Results on long datasets: Pubmed, ArXiv, and Billsum. Here we compare the limited resource (PEG$_\textrm{few\_shot}$) and full-data (PEG$_\textrm{full\_data}$) approaches with our different implementations. ($\dagger$) means that the differences are statistically significant with respect to the baseline with a p-value < 0.05 over a bootstrap hypothesis test. Best ROUGE-1/2/L scores are bolded.}
\label{tab:improvements - large datasets}
\end{table*}

\begin{figure*}[!ht]
	\centering
	\includegraphics[width=\linewidth]{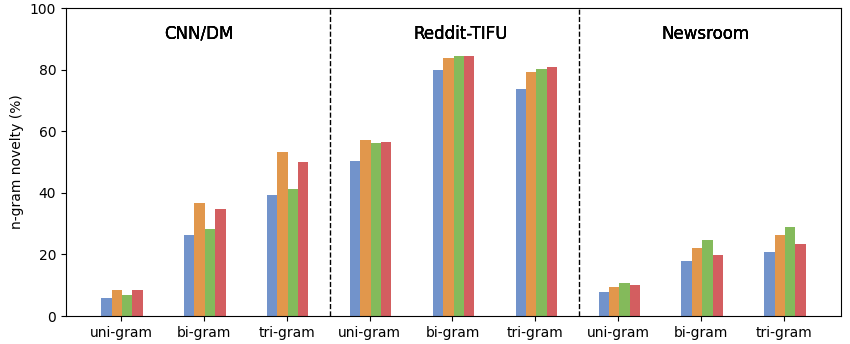}
	\caption{\label{fig:mid_ngram} Comparing the uni-, bi-, and tri-gram novelty for the medium sized datasets. These datasets contain generated sequences up to 128 tokens in length. The methods are as follows: \textcolor{blue}{\textbf{NLL (baseline)}}, \textcolor{orange}{\textbf{\textit{RwB-Hinge}}}, \textcolor{green(munsell)}{\textbf{\textit{RISK-2}}}, and \textcolor{red}{\textbf{\textit{RISK-3}}}. The unique average n-gram novelty (n-grams that do not appear in the source text) is shown to increase across the board compared to the standard NLL baseline.}
\end{figure*}

\section{Results}
\label{sec:Results}
Tables \ref{tab:improvements - small datasets}, \ref{tab:improvements - medium datasets}, and \ref{tab:improvements - large datasets} show the results of each method in comparison to the NLL-tuned baseline for the nine reported datasets. Each table reports the few-shot (top halves) and full-data results (bottom halves), where the scores have been averaged over three independently-initialised training runs. Each fine-tuning method is employed in a mixed loss framework, as mentioned in (\ref{eq:mixed1}) in Section \ref{subsec:train_obj}; the value for the $\gamma$ hyperparameter has been determined over the validation set as described in Appendix \ref{app:val_scores}. The results show that all the fine-tuning methods have surpassed the NLL  baselines for almost all datasets. Several of these improvements have also passed a bootstrap test for statistical significance, which is regarded as a more appropriate statistical test for summarisation compared to a $t$-test \citep{dror2018}.

Figure \ref{fig:mid_ngram} compares the effect that each fine-tuning method has had over the production of novel n-grams during test time (a property nicknamed as \textit{n-gram novelty}). For medium sized datasets in particular, the reinforcement learning approaches appear to, on average, facilitate the production of more distinct uni-, bi-, and tri-grams at test time, compared to the NLL baseline. Whilst n-gram novelty is typically used in summarisation to showcase test-time summary abstractiveness, the results in Figure \ref{fig:mid_ngram} highlight that training with objectives that promote sample variation leads to models capable of producing more novel n-grams (up to 13.8 pp in tri-gram novelty over CNN/DM). This is supported by the qualitative example in Table \ref{tab:cnndm_examples} which shows that the proposed fine-tuning methods can achieve greater diversity of summary predictions,
whilst still improving over the baseline NLL ROUGE scores. It seems that the proposed fine-tuning methods have allowed the model to effectively weigh the predicted summaries during training, and when combined with the ``stable'' NLL in a mixed-loss approach, this has been able to produce well-rounded predictions, diverse enough to stray from the original baseline and the reference summaries.

\begin{figure}[!ht]
	\centering
	\includegraphics[width=\linewidth]{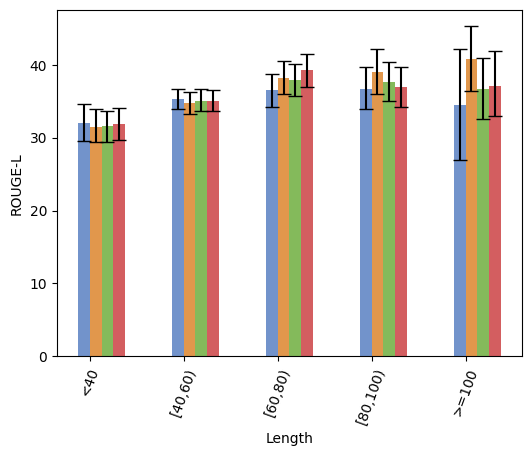}
	\caption{\label{fig:rlsum-cnndmlength} Comparison of each method for the full-data approach over a medium size dataset (CNN/DM). The methods are as follows: \textcolor{blue}{\textbf{NLL (baseline)}}, \textcolor{orange}{\textbf{\textit{RwB-Hinge}}}, \textcolor{green(munsell)}{\textbf{\textit{RISK-2}}}, and \textcolor{red}{\textbf{\textit{RISK-3}}}. We see that the reinforcement learning approaches have led, on average, to higher ROUGE-L scores for the longer summaries compared to the NLL baseline.}
\end{figure}

In addition, Figure \ref{fig:rlsum-cnndmlength} shows a performance comparison with respect to the length of the reference summaries for the full-data approach over a medium size dataset (CNN/DM). We see that our fine-tuning methods have led, on average, to higher ROUGE-L scores for the longer summaries (up to 2.3 ROUGE-L points for summaries between 80-100 tokens, and up to 6.2 points for summaries over 100 tokens). Likely, the proposed methods have been able to amend the reported tendency of the NLL models to curtail the prediction of long summaries. 

\begin{table*}[!ht]
\small
    \begin{tabular}{p{0.97\textwidth}}
        \toprule
\textbf{Source Document} \\\midrule
Dougie Freedman is on the verge of agreeing a new two-year deal to remain at Nottingham Forest. Freedman has stabilised Forest since he replaced cult hero Stuart Pearce and the club's owners are pleased with the job he has done at the City Ground. Dougie Freedman is set to sign a new deal at Nottingham Forest. Freedman has impressed at the City Ground since replacing Stuart Pearce in February. They made an audacious attempt on the play-off places when Freedman replaced Pearce but have tailed off in recent weeks. That has not prevented Forest's ownership making moves to secure Freedman on a contract for the next two seasons. \\\midrule
\textbf{Reference} \\\midrule
Nottingham Forest are close to extending Dougie Freedman's contract. The Forest boss took over from former manager Stuart Pearce in February. Freedman has since lead the club to ninth in the Championship. \\\midrule
\textbf{NLL} (40.00/30.43/32.85) \\\midrule
Dougie Freedman set to sign new deal at Nottingham Forest. Freedman has stabilised Forest since he replaced Stuart Pearce. Forest's owners are pleased with Freedman's job. \\\midrule
\textbf{\textit{RwB-Hinge}} (49.00/36.24/34.43) \\\midrule
Dougie Freedman \textcolor{blue}{\textbf{is}} set to sign \textcolor{blue}{\textbf{a}} new \textcolor{blue}{\textbf{two-year}} deal at Nottingham Forest. \textcolor{blue}{\textbf{The City Ground boss}} has stabilised \textcolor{blue}{\textbf{the club}} since he replaced Stuart Pearce. Forest's owners are pleased with Freedman's job at \textcolor{blue}{\textbf{the club}}. \\\midrule
\textbf{\textit{RISK-2}} (50.66/44.59/44.00) \\\midrule
Dougie Freedman set to sign \textcolor{blue}{\textbf{a}} new \textcolor{blue}{\textbf{two-year}} deal at Nottingham Forest. Freedman has stabilised Forest since he replaced Stuart Pearce \textcolor{blue}{\textbf{in February}}. Forest \textcolor{blue}{\textbf{made an audacious attempt}} at \textcolor{blue}{\textbf{the play-off places when}} Freedman replaced Pearce. \\\midrule
\textbf{\textit{RISK-3}} (49.33/40.54/40.00) \\\midrule
Dougie Freedman set to sign new deal at Nottingham Forest. Freedman has stabilised \textcolor{blue}{\textbf{the club}} since he replaced Stuart Pearce \textcolor{blue}{\textbf{in February}}. \textcolor{blue}{\textbf{The club's}} owners are pleased with \textcolor{blue}{\textbf{the}} job Freedman has \textcolor{blue}{\textbf{done}} at \textcolor{blue}{\textbf{the City Ground}}. \\
        \bottomrule
    \end{tabular}
    \caption{Example of the performance of each method from the CNN/DailyMail dataset for the full-data approach, compared to the reference summary and NLL baseline. Words highlighted in \textcolor{blue}{\textbf{blue}} indicate that they are not present in the baseline NLL summary. Here we choose a typical method that aligns the best with the average NLL baseline score, and compare how the methods pit against it. We see that there is a relative increase in ROUGE scores, whilst diversifying the output.}
    \label{tab:cnndm_examples}
\end{table*}

\begin{table*}[!ht]
\small
\renewcommand\multirowsetup{\centering}
\resizebox{\textwidth}{!}{%
\begin{tabular}{|c|c|c|c|c|}
\hline
\textbf{Dataset} & \textbf{Approach} & \textbf{RwB-Hinge} & \textbf{RISK-2} & \textbf{RISK-3} \\
\hline
\multirow{2}{100pt}{XSum (short)} & Few-Shot & 43.90/20.18/35.59 & \textbf{44.03}/20.28/35.75 & 43.80/\textbf{20.30}/\textbf{35.76} \\
 & Full-Data & 42.97/19.45/34.73 & 42.92/\textbf{19.53}/34.73 & \textbf{43.23}/19.25/\textbf{35.06} \\
\hline
\multirow{2}{100pt}{Newsroom (medium)} & Few-Shot & 35.47/22.31/31.11 & \textbf{36.20}/\textbf{23.11}/\textbf{31.81} & 35.96/22.87/31.62 \\
 & Full-Data & \textbf{38.17}/\textbf{25.37}/\textbf{34.12} & 37.02/24.36/33.21 & 37.08/25.11/33.22 \\
\hline
\multirow{2}{100pt}{Billsum (long)} & Few-Shot & 49.08/\textbf{29.96}/\textbf{37.63} & 48.19/28.84/36.68 & \textbf{49.23}/29.62/37.06 \\
 & Full-Data & \textbf{54.48}/\textbf{36.49}/\textbf{43.43} & 53.51/35.24/42.49 & 54.10/35.39/42.50 \\
\hline
\end{tabular}%
}
\caption{Scores on the validation set for short, medium, and long datasets to determine the best method for each size class. \textit{RISK}, on average, appears to work best for short/medium sized datasets (up to 128 tokens), and \textit{RwB-Hinge} works better for longer datasets (over 128 tokens).}
\label{tab:best_method_validation}
\end{table*}

\begin{table*}[!ht]
\centering
\renewcommand\multirowsetup{\centering}
\begin{tabular}{|c|c|c|}
\hline
\textbf{Dataset} & \textbf{RwB: No Hinge-Loss} & \textbf{RwB: with Hinge-Loss} \\
\hline
\multirow{1}{125pt}{XSum (short)} & 42.82/19.32/34.43 & \textbf{42.97}/\textbf{19.45}/\textbf{34.73} \\
\hline
\multirow{1}{125pt}{Newsroom (medium)} & \textbf{38.97}/\textbf{26.38}/\textbf{35.00} & 38.17/25.37/34.12 \\
\hline
\multirow{1}{125pt}{Billsum (long)} & 53.04/34.87/42.14 & \textbf{54.48}/\textbf{36.49}/\textbf{43.43} \\
\hline
\end{tabular}%
\caption{Comparisons between REINFORCE with baseline with and without the hinge-loss modification on the validation set for short, medium, and long datasets, to validate the use of the hinge-loss modification in our method. This is run over the full-data baselines, and shows that for the majority of dataset classes, the adopted hinge-loss modification leads to improvements in performance.}
\label{tab:rwb_comparisons}
\end{table*}

Comparing multiple fine-tuning methods is useful for showcasing the improvements that reinforcement learning can play on a generation task like summarisation. However, no single method has outperformed all others over all the datasets and in both the few-shot and full-data approaches.
Whilst all methods have achieved interesting improvements over the baseline figures, we have run a comparison over the validation set to see if their relative rankings could be a reliable indicator of the relative rankings of the test set scores reported in Tables \ref{tab:improvements - small datasets}, \ref{tab:improvements - medium datasets}, and \ref{tab:improvements - large datasets}. Table \ref{tab:best_method_validation} shows the results for one dataset per class size, showing that for the short and medium size datasets ($\leq$ 128 tokens), either of the \textit{RISK} methods could be chosen to fine-tune the model. This contrasts to the longer datasets where the hinge-loss modification has achieved the best results. In both cases, the results are in good agreement with those on the test sets.

Lastly, in Table \ref{tab:rwb_comparisons}, we further validate our use of the hinge-loss adaptation to the classical REINFORCE with baseline method -- a staple in the reinforcement learning literature of language generation tasks \citep{paulus2017deep}. Over the same three datasets of Table \ref{tab:best_method_validation}, we see that in the majority of instances the hinge-loss modification has been distinctively better than the standard approach. This confirms our intuition that the adoption of a hinge loss to restrict the gradient updates to ``good'' predictions only is beneficial to the improvement of ROUGE scores.

\section{Conclusion}
\label{sec:Conclusion}
In this paper, we have proposed two variants to the reinforcement learning approaches typically used in sequence-to-sequence learning tasks. The two proposed approaches -- nicknamed RwB-Hinge and RISK -- have been designed to improve the reinforcement learning rewards by selecting and diversifying the predictions used during the fine-tuning of the model. In a set of automated summarisation experiments over nine, diverse datasets, the approaches have consistently led to improved performance, and also diversified the generated summaries. We note that, despite its commonplace use for  summarisation evaluation, utilizing ROUGE as reinforcement learning reward does not easily translate into improved performance. For this reason, in the near future we plan to explore other contemporary score functions, such as BERTScore \citep{zhang2020bertscore}, in an attempt to build more effective rewards.

\bibliography{acl2021}
\bibliographystyle{acl_natbib}

\appendix
\onecolumn

\section{Validation Scores}
\label{app:val_scores}
To determine an appropriate $\gamma$ term for our mixed loss implementation, we have run tests with different values over the validation set for each dataset. To determine the best value, we have utilised the standard REINFORCE \citep{williams92reinforce} approach combined linearly with the negative log-likelihood. We have chosen to optimise REINFORCE here since, being a close relative, but not the same as the algorithms we have used during training, it may help to eschew overfitting. In the interest of time, we have utilised the validation scores of a single seed to determine the $\gamma$ values.\\
For the few-shot implementation in Table A.1, we have fixed the number of examples to fine-tune on (1,000) and the number of training iterations (2,000) exactly as in the standard baseline approach defined in Section \ref{sec:Experiments}. For the full-data approach in Table A.3, we have utilised all the training data, but, again in the interest of time, we have capped the number of training iterations to either: a) the same training time as the exhausted NLL tests reported in Table B.2, or b) 10,000 training iterations if the NLL training time exceeded 15,000 training iterations. 

Tables A.2 and A.4 show the best $\gamma$ values from the validation runs for all datasets. For datasets where there was no clear winner in Tables A.1 and A.3, we have compromised over the best values (highlighted in blue).\\

\begin{table*}[!ht]
\caption*{Table A.1: Validation scores of the baseline PEGASUS model, fine-tuned on a 1000 training examples for 2000 training iterations (few-shot). Best scores are highlighted.}
\resizebox{\textwidth}{!}{%
\begin{tabular}{|c|ccccc|}
\hline
\textbf{Dataset} & \textbf{0.1} & \textbf{0.3} & \textbf{0.5} & \textbf{0.7} & \textbf{0.9} \\
\hline
\textbf{AESLC} & 28.96/13.12/28.49 & 30.26/14.55/29.49 & 31.21/15.22/30.26 & 30.46/14.65/29.70 & \textcolor{blue}{31.25}/\textcolor{blue}{15.64}/\textcolor{blue}{30.42} \\
\textbf{ArXiv} & 28.06/7.99/20.70 & 33.01/10.58/21.24 & 29.49/9.32/21.12 & \textcolor{blue}{33.46/}10.46/\textcolor{blue}{22.55} & 33.43/\textcolor{blue}{10.55}/22.26 \\
\textbf{Billsum} & 41.61/28.08/34.65 & 40.37/28.07/34.17 & 40.16/28.19/34.27 & 39.56/28.11/34.16 & \textcolor{blue}{42.64/29.36/35.73} \\
\textbf{CNN/DM} & 40.30/\textcolor{blue}{18.37}/\textcolor{blue}{28.33} & 39.47/17.41/27.79 & 39.79/18.03/27.91 & 40.44/17.81/28.12 & \textcolor{blue}{40.98}/18.06/28.09 \\
\textbf{Gigaword} & 39.24/16.81/35.65 & 38.97/17.42/35.94 & 39.92/17.56/36.45 & 40.27/17.96/36.91 & \textcolor{blue}{40.91}/\textcolor{blue}{18.48}/\textcolor{blue}{37.42} \\
\textbf{Newsroom} & 36.61/25.35/33.15 & 36.93/25.39/33.25 & 36.36/24.57/32.68 & \textcolor{blue}{38.07/26.15/34.23} & 35.98/23.53/32.12 \\
\textbf{Pubmed} & 31.74/10.69/19.50 & 33.44/11.37/21.35 & 34.96/12.07/21.62 & \textcolor{blue}{37.35}/\textcolor{blue}{13.02}/22.14 & 36.57/12.99/\textcolor{blue}{22.47} \\
\textbf{Reddit-TIFU} & 19.43/4.45/15.74 & 24.87/6.56/20.08 & 25.00/6.19/19.99 & 25.73/6.85/20.55 & \textcolor{blue}{26.50}/\textcolor{blue}{6.90}/\textcolor{blue}{20.86} \\
\textbf{XSum} & 41.19/17.59/32.90 & 41.28/17.48/32.27 & 41.79/17.97/32.65 & 42.30/18.80/34.11 & \textcolor{blue}{43.43/19.58/34.76} \\
\hline
\end{tabular}%
}
\label{tab:val_1k}
\end{table*}

\begin{table*}[!ht]
\small
\caption*{Table A.2: A summary of the corresponding gamma weights determined from the above few-shot validation tests.}
\resizebox{\textwidth}{!}{%
\begin{tabular}{|c|c|c|c|c|c|c|c|c|}
\hline
\textbf{AESLC} & \textbf{ArXiv} & \textbf{Billsum} & \textbf{CNN/DM} & \textbf{Gigaword} & \textbf{Newsroom} & \textbf{Pubmed} & \textbf{Reddit-TIFU} & \textbf{XSum} \\
\hline
0.9 & 0.7 & 0.9 & 0.9 & 0.9 & 0.7 & 0.7 & 0.9 & 0.9 \\
\hline
\end{tabular}%
}
\label{tab:val_1k_gammas}
\end{table*}

\begin{table*}[!ht]
\caption*{Table A.3: Validation scores of the baseline PEGASUS model, fine-tuned on all training examples provided with the dataset for as many training iterations as either; the NLL baseline tests in Section \ref{sec:Experiments}, or 10,000 training iterations for longer datasets (ArXiv, Billsum, Pubmed). Best scores are highlighted.}
\centering
\resizebox{\textwidth}{!}{%
\begin{tabular}{|c|ccccc|}
\hline
\textbf{Dataset} & \textbf{0.1} & \textbf{0.3} & \textbf{0.5} & \textbf{0.7} & \textbf{0.9} \\
\hline
\textbf{AESLC} & 28.66/11.52/28.35 & 32.81/15.45/32.48 & 33.39/15.77/32.98 & 33.23/16.36/32.75 & \textcolor{blue}{34.94/17.17/34.11} \\
\textbf{ArXiv} & 5.71/0.00/5.56 & 1.76/0.23/1.70 & 1.61/0.04/1.59 & 10.08/1.40/9.09 & \textcolor{blue}{13.19/2.46/11.59} \\
\textbf{Billsum} & 6.50/1.50/6.45 & 9.85/4.51/9.42 & 15.50/6.31/13.04 & 32.78/17.36/25.92 & \textcolor{blue}{38.98/22.84/30.62} \\
\textbf{CNN/DM} & 3.50/0.004/0.35 & 15.37/5.75/14.91 & 24.36/8.12/22.58 & 29.17/11.46/27.44 & \textcolor{blue}{35.56/14.87/33.29} \\
\textbf{Gigaword} & 28.48/11.90/27.23 & 39.89/18.35/37.28 & 41.61/18.89/38.49 & \textcolor{blue}{43.67/20.51/40.30} & 42.68/19.34/39.26 \\
\textbf{Newsroom} & 31.48/21.03/28.32 & 27.73/15.08/24.05 & 26.78/13.79/22.84 & 33.92/20.89/30.19 & \textcolor{blue}{35.56/22.58/31.77} \\
\textbf{Pubmed} & 1.04/0.12/1.03 & 0.29/0.00/0.29 & 0.77/0.08/0.76 & 6.34/1.78/5.12 & \textcolor{blue}{10.98/2.29/8.96} \\
\textbf{Reddit-TIFU} & 1.08/0.06/1.08 & 11.59/1.43/10.45 & 9.15/1.24/8.63 & 14.71/2.58/12.59 & \textcolor{blue}{23.25/5.79/18.94} \\
\textbf{XSum} & 23.04/6.44/17.45 & 34.02/12.04/25.35 & 35.56/12.61/26.10 & 38.84/15.98/30.94 & \textcolor{blue}{41.60/18.16/33.43} \\
\hline
\end{tabular}%
}
\label{tab:val_all}
\end{table*}

\begin{table*}[!ht]
\small
\caption*{Table A.4: A summary of the corresponding gamma weights determined from the above full-data validation tests.}
\resizebox{\textwidth}{!}{%
\begin{tabular}{|c|c|c|c|c|c|c|c|c|}
\hline
\textbf{AESLC} & \textbf{ArXiv} & \textbf{Billsum} & \textbf{CNN/DM} & \textbf{Gigaword} & \textbf{Newsroom} & \textbf{Pubmed} & \textbf{Reddit-TIFU} & \textbf{XSum} \\
\hline
0.9 & 0.9 & 0.9 & 0.9 & 0.7 & 0.9 & 0.9 & 0.9 & 0.9 \\
\hline
\end{tabular}%
}
\label{tab:val_max_gammas}
\end{table*}

\newpage
\section{Model Hyperparameters}
\label{app:hyperparameters}
In our experiments, we have utilised the same hyperparameters used in the original PEGASUS paper \citep{zhang2019}. The exception to this is our use of a smaller batch size, constrained by computational resources. As batch size we have used 1, which has resulted in a drop in performance compared to that of the original paper. However, our fine-tuning approach is ensured to converge through the use of a convergence criterion. This is defined by a validation run that evaluates the model every 1000 training iterations, and monitors the progression of the validation loss over the entire training run. A model is deemed `converged' if its validation loss does not decrease over 3000 training iterations.

\begin{table*}[!ht]
\caption*{Table B.1: Model hyperparameters used in the few-shot experiments. All values except the fine-tuning steps are also used in the full-data approach.}
\centering
\resizebox{\textwidth}{!}{%
\begin{tabular}{|c|ccccccc|}
\hline
\textbf{Dataset} & \textbf{Learning Rate} & \textbf{Label Smoothing} & \textbf{Fine-Tuning Steps} & \textbf{Batch Size} & \textbf{Beam Size} & \textbf{Max Input Tokens} & \textbf{Max Target Tokens} \\
\hline
\textbf{AESLC} & 5e-4 & 0.1 & 2000 & 1 & 1 & 512 & 32 \\
\textbf{ArXiv} & 5e-4 & 0.1 & 2000 & 1 & 1 & 1024 & 256 \\
\textbf{Billsum} & 5e-4 & 0.1 & 2000 & 1 & 1 & 1024 & 256 \\
\textbf{CNN/DM} & 5e-4 & 0.1 & 2000 & 1 & 1 & 1024 & 128 \\
\textbf{Gigaword} & 5e-4 & 0.1 & 2000 & 1 & 1 & 128 & 32 \\
\textbf{Newsroom} & 5e-4 & 0.1 & 2000 & 1 & 1 & 512 & 128 \\
\textbf{Pubmed} & 5e-4 & 0.1 & 2000 & 1 & 1 & 1024 & 256 \\
\textbf{Reddit-TIFU} & 5e-4 & 0.1 & 2000 & 1 & 1 & 1024 & 128 \\
\textbf{XSum} & 5e-4 & 0.1 & 2000 & 1 & 1 & 512 & 64 \\
\hline
\end{tabular}%
}
\label{tab:low_resource_hps}
\end{table*}

\begin{table*}[!ht]
\caption*{Table B.2: Model fine-tuning steps used in the full-data experiments. The NLL and all fine-tuning tests (except the validation tests), were validated every 1000 training iterations on a separate validation set, with the validation loss monitored over the run. An early stopping criterion was in place to stop training if the validation loss had not declined in 3000 consecutive training iterations. All methods have been averaged over three seed runs, whereas for the validation run we report results from a single run.}
\centering
\begin{tabular}{|c|cc:ccc|}
\hline
\textbf{Dataset} & \textbf{NLL} & \textbf{Validation} & \textbf{RwB-Hinge} & \textbf{RISK-2} & \textbf{RISK-3} \\
\hline
\textbf{AESLC} & 7k & 7k & 5k & 5.3k & 5.3k \\
\textbf{ArXiv} & 43k & 10k & 7k & 7k & 7k \\
\textbf{Billsum} & 44k & 10k & 5k & 5k & 4.6k \\
\textbf{CNN/DM} & 12k & 12k & 6.6k & 6.6k & 7.6k \\
\textbf{Gigaword} & 10k & 10k & 5.6k & 6k & 6k \\
\textbf{Newsroom} & 10k & 10k & 6.3k & 6.6k & 6.3k \\
\textbf{Pubmed} & 55k & 10k & 5.6k & 6k & 6k \\
\textbf{Reddit-TIFU} & 10k & 10k & 7k & 7k & 6.5k \\
\textbf{XSum} & 8k & 8k & 6k & 5.3k & 6k \\
\hline
\end{tabular}%
\label{tab:max_hps}
\end{table*}

\end{document}